\def\year{2021}\relax
\documentclass[letterpaper]{article} 
\usepackage{aaai21}  
\usepackage{times}  
\usepackage{helvet} 
\usepackage{courier}  
\usepackage[hyphens]{url}  
\usepackage{graphicx} 
\usepackage{natbib}  
\usepackage{caption} 
\DeclareGraphicsExtensions{.png}
\graphicspath{{pic/}}

\usepackage{amsmath, algcompatible, algorithm, amssymb, url, multirow, enumerate, cite, diagbox, makecell, subfigure}
\usepackage[switch]{lineno}
\newcommand{\BL}{\mathcal{L}}
\newcommand{\mathcalX}{\mathcal{X}}
\newcommand{\mathcalY}{\mathcal{Y}}
\newcommand{\mathbbE}{\mathbb{E}}

\title{Temporal Pyramid Network for Pedestrian Trajectory Prediction with Multi-Supervision}

\author{
   	Rongqin Liang,\textsuperscript{\rm 1, 2}
    Yuanman Li,\textsuperscript{\rm 1, 2,}\thanks{The corresponding author.}
    Xia Li,\textsuperscript{\rm 1, 2}
    yi tang,\textsuperscript{\rm 1, 2}
    Jiantao Zhou,\textsuperscript{\rm 3, 4}
    Wenbin Zou\textsuperscript{\rm 1, 2} \\
}
\affiliations{

	\textsuperscript{\rm 1} College of Electronics and Information Engineering, Shenzhen University\\
	\textsuperscript{\rm 2} Guangdong Key Laboratory of Intelligent Information Processing\\
	\textsuperscript{\rm 3} Department of Computer and Information Science, University of Macau\\
	\textsuperscript{\rm 4} State Key Laboratory of Internet of Things for Smart City\\

    1810262064@email.szu.edu.cn,
    \{lixia, yuanmanli, yitang, wzou\}@szu.edu.cn,
    jtzhou@um.edu.mo
}

\nocopyright

\begin{document}

\maketitle
\begin{abstract}
Predicting human motion behavior in a crowd is important for many applications, ranging from the natural navigation of autonomous vehicles to intelligent security systems of video surveillance.  
All the previous works model and predict the trajectory with a single resolution, which is rather inefficient and difficult to simultaneously exploit the long-range information (e.g., the destination of the trajectory), and the short-range information (e.g., the walking direction and speed at a certain time) of the motion behavior. 
In this paper, we propose a temporal pyramid network for pedestrian trajectory prediction through a squeeze modulation and a dilation modulation. Our hierarchical framework builds a feature pyramid with increasingly richer temporal information from top to bottom, which can better capture the motion behavior at various tempos. Furthermore, we propose a coarse-to-fine fusion strategy with multi-supervision. By progressively merging the top coarse features of global context to the bottom fine features of rich local context, our method can fully exploit both the long-range and short-range information of the trajectory. 
Experimental results on several benchmarks demonstrate the superiority of our method. 
\end{abstract}

\section{Introduction}
Modeling the behaviors of pedestrians is an essential step for many applications, including self-driving platforms for safe decision making \cite{liang2019peeking}, socially-aware robots for natural navigation \cite{monfort2015intent} and surveillance systems to identify suspicious activities \cite{BastaniTIP2016}. Trajectory prediction as one of the most important future behavior modeling tasks, aims to predict possible future trajectories according to historical paths in the last few seconds \cite{alahi2016social, gupta2018social, mohamed2020social,xu2020AAAI}. 
Despite its importance, predicting the trajectory is very challenging due to the inherent properties of pedestrians. First, human motions are highly \textit{multimodal}, which means that there could be several socially-acceptable and distinct future behaviors under the same trajectory history. 
Second, human motions are highly affected by the people around them, 
Jointly modeling the complex social behaviors is rather challenging in reality. 

Traditional pedestrian trajectory prediction algorithms heavily rely on the handcrafted rules to describe human motions \cite{helbing1995social,pellegrini2009you}, 
which are difficult to generalize in complex new scenes.
Recently, the data-driven based algorithms have received significant attention in the community. Among them, RNN and its variant LSTM have been widely adopted. Social-LSTM \cite{alahi2016social} as one of the earliest works on pedestrian trajectory prediction, encoded the motion information using a recurrent network. CIDNN \cite{xu2018encoding} considered different importance of persons to a target pedestrian in a crowd interaction module. The recent works PIF \cite{liang2019peeking} and  SR-LSTM \cite{zhang2019sr} enhanced the performance of Social-LSTM by taking the scene context as side information. Though the RNN architecture endowed above methods to learn and predict trajectories in a data-driven manner, they failed to capture the multimodal nature of human.

\begin{figure*}[!t]
	\centering
	\includegraphics[width=0.932\linewidth]{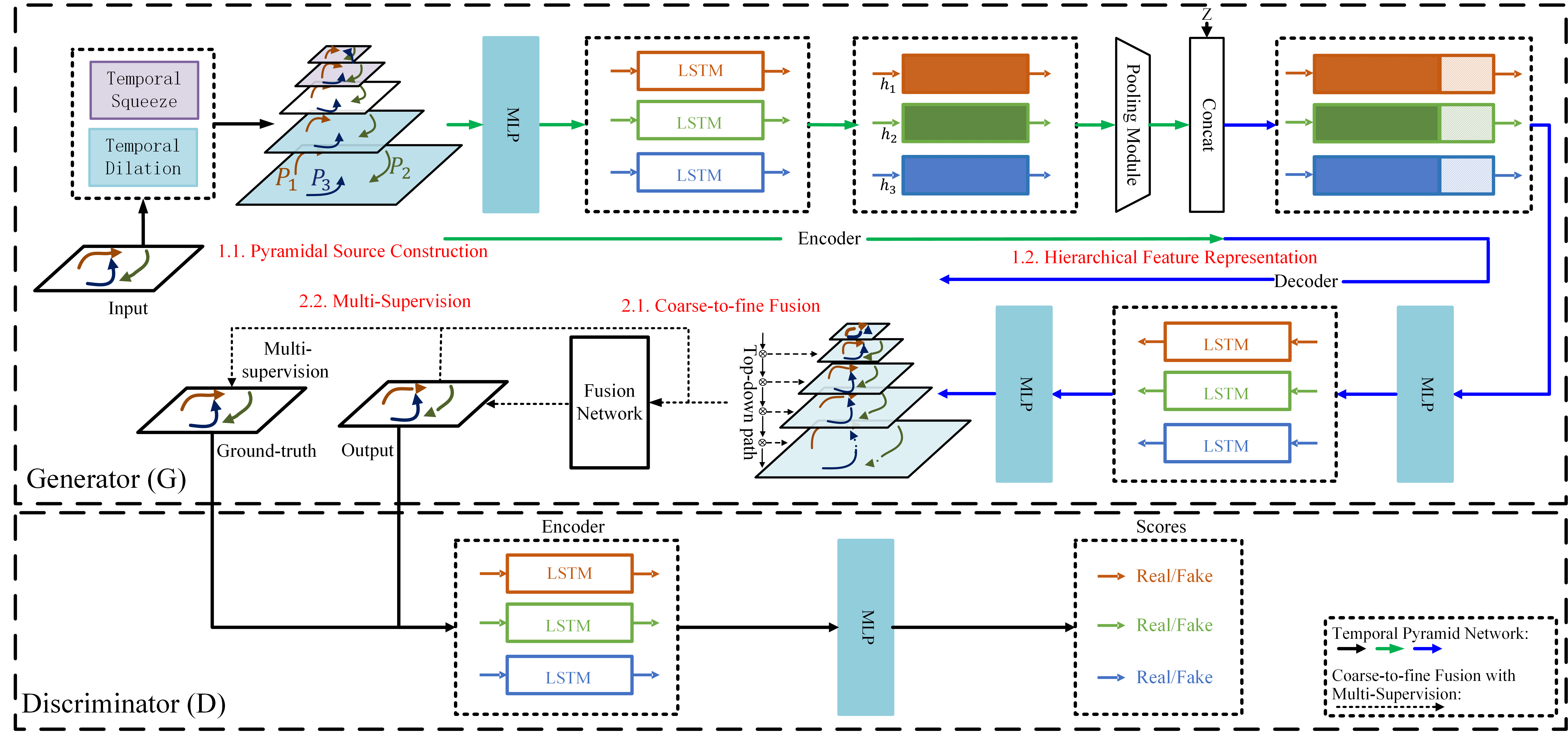}
	\caption{The framework of our proposed TPNMS. The network consists of a generator and a discriminator. The input of the generator is the historical trajectories of pedestrians, and the output is the corresponding predicted future trajectories. The pyramidal source is first constructed through the temporal squeeze modulation and the temporal dilation modulation. Then, an encoder-decoder network is adopted for hierarchical feature learning. Features are finally fed into a fusion network (presented in Fig. \ref{fig:fusion}) to generate the future trajectories with multi-supervision.}
	\label{fig:framework}
\end{figure*}

In order to produce multiple socially-acceptable trajectories, some researchers suggested constructing the recurrent models with generative settings, which led to learning the distribution of the future trajectory rather than directly generating a deterministic path \cite{gupta2018social, van2019safecritic, Li2019IROS, zhao2019multi}. Social-GAN \cite{gupta2018social} is the pioneering trajectory prediction work incorporating the LSTM model with the generative adversarial networks (GANs) \cite{goodfellow2014generative}, permitting to produce multiple plausible trajectories. SoPhie \cite{sadeghian2019sophie} improved social-GAN through a scene feature extraction component. 
Some researchers proposed to use graph to model the social interactions \cite{ ma2019trafficpredict, vemula2018social,socialbigat2019,huang2019stgat, ivanovic2019trajectron,shi2020AAAI, mohamed2020social}. 
For example, the most recent work Social-STGCNN \cite{mohamed2020social} modeled trajectories using the spatio-temporal graph convolution neural network, and achieved promising performance.

Though trajectory prediction has been studied from many aspects, all the existing methods encoded and decoded the trajectory with a single resolution (\textit{i.e.}, a fixed length of time steps). This makes them fail to fully exploit the temporal relations of the motion behavior. We argue that simultaneously modeling the global context (\textit{e.g.}, where the pedestrian plans to go) and  the local context (\textit{e.g.}, the direction and speed at a certain time) with a single resolution is inefficient or rather difficult, if possible.

To alleviate the above limitation, in this work, we propose a novel Temporal Pyramid Network with Multi-Supervision (TPNMS) for pedestrian trajectory prediction. As shown in Fig. \ref{fig:framework}, our framework consists of a generator $G$ and a discriminator $D$, which are trained in opposition to each other.
First, we devise a pyramid feature extractor composed of a squeeze module and a dilation module for multi-scale feature generation from a fixed length input trajectory. The pyramidal features are then fed into an RNN based Encoder-Decoder to generate hierarchical representations of the motion. To ensure effective representations of all pyramid levels, we further propose a coarse-to-fine fusion strategy with multi-supervision through progressively combining higher pyramid levels with lower ones. 
Finally, similar to Social-GAN, our network is trained in an adversarial manner to produce multiple socially-acceptable motion trajectories, conforming to the multimodal behavior of pedestrians.

It should be noted that most of the previous pyramid representation methods were designed in spatial domain and only for detection or recognition tasks. To the best of our knowledge, this is the \textit{first} attempt that models trajectories in a scene as temporal pyramids. As will be shown later, our method outperforms previous approaches by a big margin on several datasets. 
The main contributions of our work are:
\begin{itemize}
	\item A novel temporal pyramid network is proposed to capture the motion behaviors of pedestrians at various tempos. With our hierarchical design, both short-range and long-range motion behaviors can be effectively exploited.  
	\item By progressively combining the global context with the local one, we further propose a coarse-to-fine trajectory modeling in a multi-supervised fashion.
	\item Our hierarchical design can be regarded as auxiliary modules, and easily extended to other sequence prediction frameworks, thus bringing performance improvements.
\end{itemize}


%


\section{Proposed Temporal Pyramid Network with Multi-Supervision (TPNMS)} \label{sec:proposed}

\begin{figure*}[!t]
	\centering
	\subfigure[Temporal squeeze modulation]{
		\includegraphics[width=2.73in]{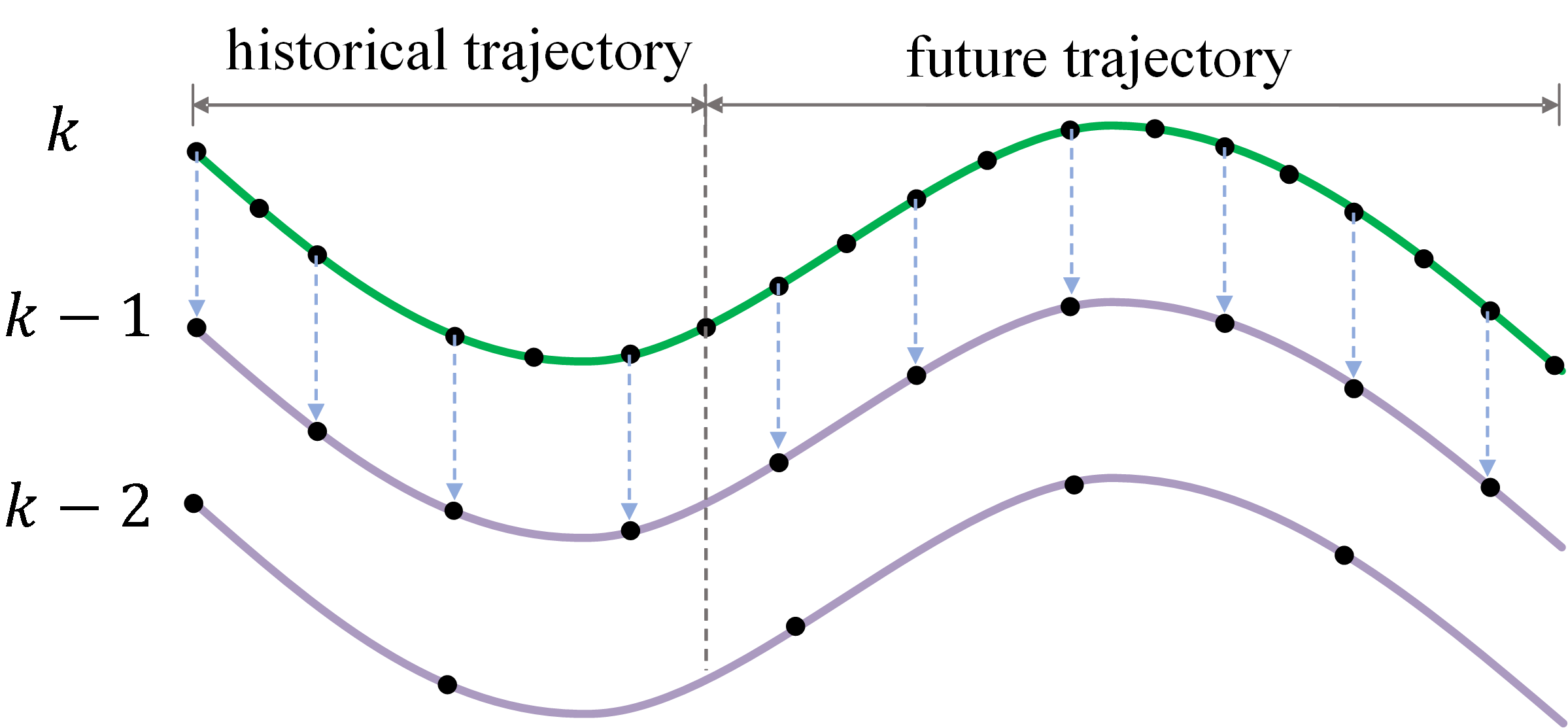}}~~~~~~~~~~
	\subfigure[Temporal dilation modulation]{
		\includegraphics[width=2.73in]{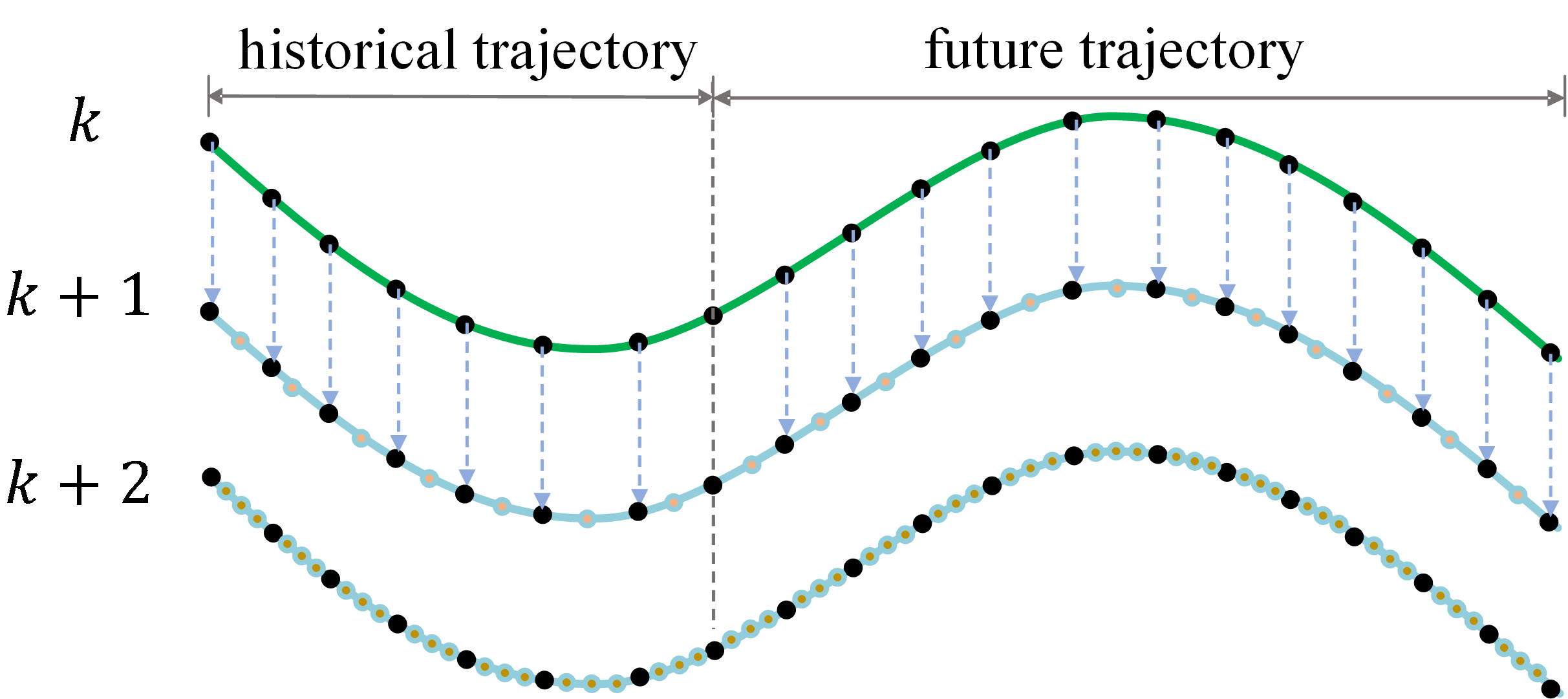}}	
	\caption{Illustration of the proposed (a) temporal squeeze modulation, and (b) temporal dilation modulation.}
	\label{fig:TPL}
\end{figure*}
\subsection{Problem Formulation}
Given a set of $N$ pedestrians in a scene with observed positions over a fixed duration, the trajectory prediction algorithm aims to jointly reason and forecast the future trajectories of all pedestrians. Let $(x_i^t, y_i^t)$ be the position of the $i$-th pedestrian at the time step $t$, where $i \in \{1,...,N\}$. Denote $X_i^{(t_1:t_2)}=[(x_i^{t_1}, y_i^{t_1}),...,(x_i^{t_2}, y_i^{t_2})]$ as the observed historical trajectory of the $i$-th pedestrian from the time step $t_1$ to $t_2$. Similarly, we define $Y_i^{(t_1:t_2)}$ as the future trajectory of the $i$-th pedestrian from the time $t_1$ to $t_2$.
The trajectory prediction algorithm takes as input the previous trajectories with $t_o$ time steps of all pedestrians in a scene, denoted by
\begin{equation}
\mathcalX = \{X_1^{(1:t_o)},...,X_N^{(1:t_o)}\},
\end{equation}
and aims to predict their trajectories in the next $t_p$ time steps simultaneously. We use $\mathcalY$ to represent the true future trajectories, i.e.,
\begin{equation}
\mathcalY = \{Y_1^{(t_o+1:t_o+t_p)},...,Y_N^{(t_o+1:t_o+t_p)}\}.
\end{equation}

For the sake of brevity, we hereafter will drop the superscript when there is no ambiguity, i.e., $X_i \triangleq X_i^{(1:t_o)}$ and $Y_i \triangleq Y_i^{(t_o+1:t_o+t_p)}$. We further use $X$ and $Y$ to represent a generic historical trajectory and the corresponding future trajectory, respectively.

\subsection{TPN for Trajectory Prediction}\label{sec:TPN}
Feature pyramids play a significantly important role in the field of computer vision for recognizing objects at vastly different scales \cite{he2015spatial}. For example, the popular hand-engineered feature extractors such as SIFT \cite{Lowe04} were designed to compute features in a multi-scale space. Lin \textit{et. al.} \cite{lin2017feature} accommodated the idea of pyramid representation to deep convolutional neural networks, achieving quite promising performance in the detection task. Most of the previous approaches were designed in spatial domain. More recently, some works proposed to extract hierarchical features in temporal domain, and demonstrated its effectiveness in action recognition \cite{yang2020temporal} and scene classification \cite{huang2019attentive}.

Motivated by the great success of the pyramid representation, we propose a temporal pyramid network (TPN) tailored for pedestrian trajectory prediction. Compared with the existing algorithms which model the trajectory with a single resolution, 
our temporal pyramid architecture is effective in exploiting the motion behaviors at various tempos, and the hierarchical generation process could greatly facilitate the joint modeling of both global and local contexts. Besides, benefiting from the LSTM network, all levels of pyramids share the same parameters. This allows our method to operate on a single-branch backbone regardless of how many levels are adopted, then avoid to increase model complexity.

For better illustration, we decompose our TPN into the following two components: 1) pyramidal source construction, and 2) hierarchical feature representation. 

\subsubsection{Pyramidal source construction} \label{sec:psc}
For each trajectory of the input $\mathcalX$, we propose to generate a set of $L$ hierarchical features with multi-resolution, and then construct a feature pyramid, having increasingly richer temporal information from top to bottom. With the aid of the pyramid framework, our method can fully exploit the short-range behavior and the long-range behavior in a hierarchical way.  As depicted in Fig. \ref{fig:framework}, this process can be summarized as two procedures, i.e., 1) the temporal squeeze modulation, and 2) the temporal dilation modulation.

\noindent\textit{\textbf{Temporal squeeze modulation:}}
Assume that there are totally $L$ scales of the temporal pyramid network for each trajectory. Denote the feature of the $k$-th scale as $X_i^k$, which is identical to $X_i$. The goal of the temporal squeeze modulation is to reduce the impact of the local context, and generate a set of features with increasingly stronger global context from $\mathcalX$. 
To this end, we propose to gradually produce the top $k-1$ scales through uniformly sampling from the scale below with an interval factor $2$.   
In this work, we refer to the above process as the temporal squeeze modulation. 

Fig. \ref{fig:TPL}(a) illustrates the procedure of the temporal squeeze modulation. For the $\ell$-scale ($\ell<k$), the feature can be represented as 
\begin{equation}
X_i^\ell = [(\tilde{x}_i^{1}, \tilde{y}_i^{1}),...,(\tilde{x}_i^{m_\ell}, \tilde{y}_i^{m_\ell})],
\end{equation} where $m_\ell = \lceil t_o/2^{k-\ell} \rceil$, and
\begin{equation}
\tilde{x}_i^{j} = x_i^{1+2^{k-\ell}(j-1)}, ~~ \tilde{y}_i^{j} = y_i^{1+2^{k-\ell}(j-1)}.
\end{equation}
We can see that the detailed short-range information of the motion is gradually weakened by the temporal squeeze modulation,  which encourages the higher scales to capture more long-range motion behaviors of pedestrians.

\noindent\textit{\textbf{Temporal dilation modulation:}}
Note that the observed trajectories are usually of short duration, then the number of scales generated by the temporal squeeze modulation cannot fully capture the motion behaviors. To handle this issue, we further introduce a complementary procedure called temporal dilation modulation, which is similar to the dilated convolution operator widely used in various vision tasks.
The temporal dilation modulation could generate more dense trajectories for hierarchical feature representation, then exploit richer short-range information of the motion.

We propose to conduct the temporal dilation modulation through trajectory interpolation. It should be noted that pedestrians usually walk/run at varying speeds, accelerations and in different directions over time. In order to generate smooth dense trajectories, in this work, we adopt the cubic spline algorithm for the trajectory interpolation. For simplicity, we rewrite the observed trajectory of the $i$-th pedestrian as a series of time-position pairs
\begin{equation}
TX_i = \big[(1,(x_i^{1}, y_i^{1})),...,(t_o, (x_i^{t_o}, y_i^{t_o}))\big].
\end{equation}
We adopt the cubic spline algorithm to seek for a piecewise-cubic function $f(t)$: $\mathbb{R}\rightarrow\mathbb{R}^2$
\begin{equation}
f(t) = \begin{cases}
f_1(t), ~~~1 \leq t < 2\\
\vdots\\
f_{t_o-1}(t), ~~~t_o-1 \leq t \leq t_o \\
\end{cases},
\end{equation} where 
\begin{equation}
f_k(t) = a_k +b_k(t-k)+c_k(t-k)^2+d_k(t-k)^3
\end{equation} represents the curve between the time steps $k$ and $k+1$, and $a_k, b_k, c_k, d_k \in \mathbb{R}^2$ are parameters of the cubic spline. According to the cubic spline algorithm, given the trajectory $X_i$, there exists a unique set of parameters $\{a_k,b_k,c_k,d_k\}_{k=1,\cdots,t_o-1}$ such that the resulting trajectory curve passes through all the positions in $X_i$ with continuous velocity and acceleration at each position. 
Upon having $f(t)$, the feature $X_i^{\ell}$ at the $\ell$-scale ($\ell>k$) can be calculated by interpolating positions of the in-between and unobserved time steps as shown in Fig. \ref{fig:TPL}(b). Mathematically, we have
\begin{equation}
\small X_i^{\ell} = [f(1), f(1+\frac{1}{c}),f(1+\frac{2}{c}),...,f(2),...,f(t_o-\frac{1}{c}),f(t_o)],
\end{equation} where $c=2^{\ell-k}$. The interpolated dense trajectories offer more local information for the lower scales, permitting to capture more short-range motion behaviors of pedestrians.  With above two modulations, we finally construct the pyramidal source as shown in Fig. \ref{fig:framework}.

\subsubsection{Hierarchical feature representation}
For simplicity, we use a similar network architecture proposed in \cite{gupta2018social} as the backbone to extract hierarchical features from the constructed pyramid. As shown in Fig. \ref{fig:framework}, the backbone network consists of two components, \textit{i.e.}, the encoder and the decoder. 
At the encoder side, we embed the position of each pedestrian as
\begin{equation}
e_i^t = MLP(x_i^t, y_i^t; \Theta_{me}),
\end{equation} where $t\leq t_o$ and $\Theta_{me}$ represents the parameters of MLP. The embedded feature $e_i^t$ is then fed into an LSTM block, which produces the hidden state at the time step $t$
\begin{equation}\label{eq:hidden_en}
h_i^t = LSTM(h_i^{t-1},e_i^t; \Theta_{le}).
\end{equation} Note that the parameters of LSTM ($\Theta_{le}$) are shared among all the pedestrians and the scales of the pyramid.

In order to generate multiple socially-acceptable trajectories, our model is designed under the framework of GANs. 
According to GANs, at the decoder side, we concatenate a noise vector $z$ sampled from the standard normal distribution to the hidden state $h_i^{t_o}$. Further, we use the pooling module proposed by \cite{gupta2018social} to encode the influence caused by pedestrians around. We write
\begin{equation}\label{eq:hidden-to}
h_{i}^{t_o} := [(h_i^{t_o}, P_i); z].
\end{equation}
Then for each $z$, we recurrently decode the hierarchical features of the trajectory as follows
\begin{equation}\label{eq:prediction}
\begin{aligned}
\hat{e}_i^{t-1} = MLP(\hat{x}_i^{t-1},\hat{y}_i^{t-1}; \Theta_{md})\\
{h}_i^t = LSTM({h}_i^{t-1}, \hat{e}_i^{t-1}; \Theta_{ld})\\
(\hat{x}_i^t,\hat{y}_i^t) = MLP({h}_i^t; \Theta'_{md})
\end{aligned},
\end{equation} where $t\geq t_o+1$, $(\hat{x}_i^{t_o},\hat{y}_i^{t_o}) =(x_i^{t_o}, y_i^{t_o})$, and $\Theta_{md}, \Theta_{ld},\Theta'_{md}$ are parameters to be learned.

\begin{figure}[!t]
	\centering
	\includegraphics[width=3.3in]{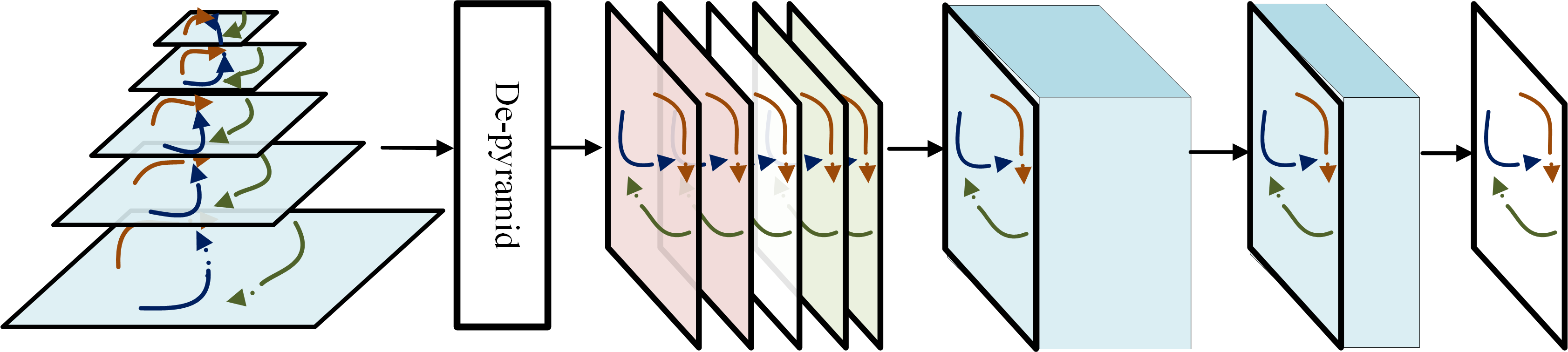}
	\caption{The framework of the fusion network, where the convolutional layers have the $1\times 1$ kernel size, and the number of channels are $8,4$ and $1$, respectively.}	\label{fig:fusion}
\end{figure} 

\begin{table*}[!t]
	\centering  
	\begin{tabular}{c|cc|cc|cc|cc|cc|cc}  
		\hline  
		\multirow{2}{*}{\diagbox{Methods}{Datasets}}&\multicolumn{2}{c|}{ETH}&\multicolumn{2}{c|}{Hotel}&\multicolumn{2}{c|}{Univ}&\multicolumn{2}{c|}{Zara1}&\multicolumn{2}{c|}{Zara2}&\multicolumn{2}{c}{AVG}\cr\cline{2-13}&ADE&FDE&ADE&FDE&ADE&FDE&ADE&FDE&ADE&FDE&ADE&FDE\cr
		\hline
		Linear&1.33&2.94&0.39&0.72&0.82&1.59&0.62&1.21&0.77&1.48&0.79&1.59\cr
		S-LSTM&1.09&2.35&0.79&1.76&0.67&1.40&0.47&1.00&0.56&1.17&0.72&1.54\cr
		S-GAN&0.81&1.52&0.72&1.61&0.60&1.26&0.34&0.69&0.42&0.84&0.58&1.18\cr 
		S-GAN-P&0.87&1.62&0.67&1.37&0.76&1.52&0.35&0.68&0.42&0.84&0.61&1.21\cr
		PIF&0.73&1.65&0.30&0.59&0.60&1.27&0.38&0.81&0.31&0.68&0.46&1.00\cr
		SoPhie&0.70&1.43&0.76&1.67&0.54&1.24&\textbf{0.30}&0.63&0.38&0.78&0.54&1.15\cr
		STGAT &0.65&1.12&0.35&0.66&0.52&1.10&0.34&0.69&0.29&0.60&0.43&0.83\cr
		SR-LSTM &0.63&1.25&0.37&0.74&0.51&1.10&0.41&0.90&0.32&0.70&0.45&0.94\cr
		Social-BiGAT &0.69&1.29&0.49&1.01&0.55&1.32&\textbf{0.30}&0.62&0.36&0.75&0.48&1.00\cr
		Social-STGCNN&0.64&1.11&0.49&0.85&\bf0.44&\bf0.79&0.34&\bf0.53&0.30&\bf0.48&0.44&0.75\cr 
		\bf TPNMS&\bf0.52&\bf0.89&\bf0.22&\bf0.39&0.55&1.13&0.35&0.70&\bf0.27&0.56&\bf0.38&\bf0.73\cr
		\hline
	\end{tabular}
	\caption{The performance of different methods in terms of ADE / FDE metrics.}
	\label{tb:comparison}
\end{table*}

\subsection{Coarse-to-fine Fusion with Multi-supervision}
To merge and exploit the information of hierarchical features generated by the above TPN, we further propose a coarse-to-fine fusion strategy with multi-supervision.
\subsubsection{Coarse-to-fine fusion}
As shown in Fig. \ref{fig:framework}, the coarse-to-fine fusion organizes features in a top-down pathway, where the top coarse features with long-range context are progressively merged to the bottom fine features with rich local-range context.  Denote the extracted feature of the $\ell$-scale as $\hat{X}_i^\ell$, which is updated by merging the information of the above scale. We write
\begin{equation}
\hat{X}_i^\ell := \frac{1}{2}(\hat{X}_i^\ell\oplus\hat{X}_{i, \uparrow}^{\ell-1}),
\end{equation} where $\hat{X}_{i, \uparrow}^{\ell-1}$ means upsampling $\hat{X}_{i}^{\ell-1}$ by a factor of 2, and $\oplus$ serves as the element-wise addition. This process is iterated until the finest resolution feature is merged. With the coarse-to-fine fusion strategy, the long-range and local-range motion information is collaborated, and then can complete to each other.

\subsubsection{Multi-supervision}
To ensure effective representation of each pyramid level, all the scales are supervised during training, where the corresponding loss function is formulated as 
\begin{equation}
\BL_s = \frac{1}{NL}\sum_{i=1}^{N}\sum_{\ell=1}^{L}\lambda_\ell\|\hat{X}_i^\ell - Y_i^\ell \|_2^2.
\end{equation} 
Here $Y_i^\ell$ is the ground-truth pyramidal source of the future trajectory, which can be constructed from $Y_i$ in the same way detailed in Section \ref{sec:psc}. The hyper-parameter $\lambda_\ell$ is inversely proportional to the feature length of $\hat{X}_i^\ell$, which we empirically set
\begin{equation}
\lambda_\ell = \frac{t_p}{m'_\ell}.
\end{equation}Here $m'_\ell$ represents the length of $\hat{X}_i^\ell$.

The final predicted trajectory is produced by a fusion layer as shown in Fig. \ref{fig:fusion}, where a de-pyramid layer is first adopted to down-sample or up-sample the hierarchical features $\{\hat{X}_i^\ell\}_{\ell=1}^L$ to a fixed length $t_p$. The results are then concatenated as a tensor of size $L\times2\times t_p$, which is further processed through three convolutional layers to fuse information across the whole pyramid. The fusion layer finally generates the predicted trajectory $\hat{Y}_i$. We supervise the final output using the loss
\begin{equation}
\BL_f =\frac{1}{N} \sum_{i=1}^{N} \|\hat{Y}_i - Y_i\|_2^2.
\end{equation}

\subsubsection{Adversarial training}
The architecture of the discriminator is shown in Fig. \ref{fig:framework}, which consists of an LSTM component and an MLP component. 
The LSTM component takes as input the ground-truth trajectory $[X, Y]$ or the generated trajectory $[X, \hat{Y}]$. The last hidden state of the LSTM is fed into the MLP, which outputs the classification score. Let $\hat{Y} = G(z, X)$. The adversarial loss is defined as
\begin{equation}
\begin{aligned}
&\BL_{avd} = \mathbbE_{X,Y \sim P_{data}(X,Y)}[\log D(X,Y)] \\
& + \mathbbE_{X \sim P_{data}(X), z\sim P_z(z)}[\log (1-D(X,G(z,X))].
\end{aligned}
\end{equation}
Finally, training the network is cast into a two-player min-max game with the following objective function
\begin{equation} \label{eq:final_loss}
\min_G\max_D \BL_{avd}+\BL_s +\BL_f.
\end{equation} Above problem can be solved by alternatively updating the generator $G$ and the discriminator $D$.

\section{Experimental Results}  \label{sec:experiment}
We implement our model TPNMS using the PyTorch framework with an NVIDIA TITAN Xp GPU.
All the source code and models will be publicly available upon the acceptance. 

\subsection{Implementation Details}
The number of pyramid scales is empirically set as $5$, and the dimensions of the hidden state for the encoder and the decoder are 32. Each input coordinate $(x,y)$ is embedded as a 16-dimensional vector. The length of the noise vector $z$ is $8$. We adopt Adam algorithm \cite{kingma2014adam} to optimize the loss function (\ref{eq:final_loss}) and train our network with the following hyper-parameter settings: batch size is 64; learning rates for the Generator and Discriminator are set to be 1e-4 and 2e-4, respectively; betas are 0.9 and 0.999; weight decay is 1e-4 and the number of epochs is 400. 
\subsection{Datasets and Metrics}
\textit{Datasets}: 
We evaluate our method on two public datasets, \textit{i.e.}, ETH\cite{pellegrini2009you} and UCY\cite{lerner2007crowds}. 
These datasets consist of 5 unique scenes: ETH, HOTEL, UNIV, ZARA1 and ZARA2 with 4 different scenes. There are totally 1536 pedestrians with thousands of trajectories containing challenging behaviors such as walking together, crossing each other, forming groups and dispersing. 


\textit{Metrics}: For the sake of fairness, we use the widely adopted leave-one-out approach evaluation methodology. 
The number of observed time steps is 8 (3.2 seconds) of each person and the upcoming trajectory of 12 time steps (4.8 seconds) is used to predict.
Following prior works, we use two error metrics to evaluate the performance of different pedestrian trajectory prediction models. 
\begin{enumerate}
	\item \textit{Average Displacement Error} (ADE): The average Euclidean distance between the ground-truth trajectory and the predicted one,\\
	\begin{equation}
	 \mathrm{ADE}=\frac{\sum_{i=1}^N \sum_{t=t_o+1}^{t_o+t_p}\left\|Y_{i}^{(t)}-\hat{Y}_{i}^{(t)}\right\|_{2}}{N\times t_p}.
	\label{eq:ADE}
	\end{equation}	
	\item \textit{Final Displacement Error} (FDE): The Euclidean distance between the ground-truth destination and the predicted one,\\
	\begin{equation}
	 \mathrm{FDE}=\frac{\sum_{i=1}^N\left\|Y_{i}^{(t_o+t_p)}-\hat{Y}_{i}^{(t_o+t_p)}\right\|_2}{N}.
	\label{eq:FDE}
	\end{equation}
\end{enumerate}

\subsection{Baselines}
We compare our method \textit{\textbf{TPNMS}} with following approaches:
\textit{\textbf{Linear}} \cite{alahi2016social}: a linear regressor that predicts the next coordinates based on previous points.
\textit{\textbf{S-LSTM}} \cite{alahi2016social}: a method based on LSTM and social pooling.
\textit{\textbf{S-GAN}} and \textit{\textbf{S-GAN-P}} \cite{gupta2018social}: a model that employs GAN to generate multimodal pedestrian trajectories and the latter with a global pooling module.
\textit{\textbf{PIF}} \cite{liang2019peeking}: a multi-task method using both visual features and interaction information.
\textit{\textbf{SoPhie}} \cite{sadeghian2019sophie}: an improved GAN based model considering the physical constraints.
\textit{\textbf{SR-LSTM}} \cite{zhang2019sr}: a state refinement method for LSTM based pedestrian trajectory prediction.
\textit{\textbf{Social-BiGAT}} \cite{socialbigat2019} and \textit{\textbf{STGAT}} \cite{huang2019stgat}: methods based on GAN and graph attention.
\textit{\textbf{Social-STGCNN}} \cite{mohamed2020social}: an approach that models the social behavior of pedestrians using a graph. 
Similar to previous works, we generate 20 samples based on the predicted distribution.
\begin{figure*}[!t]
	\centering
	\includegraphics[width=\linewidth]{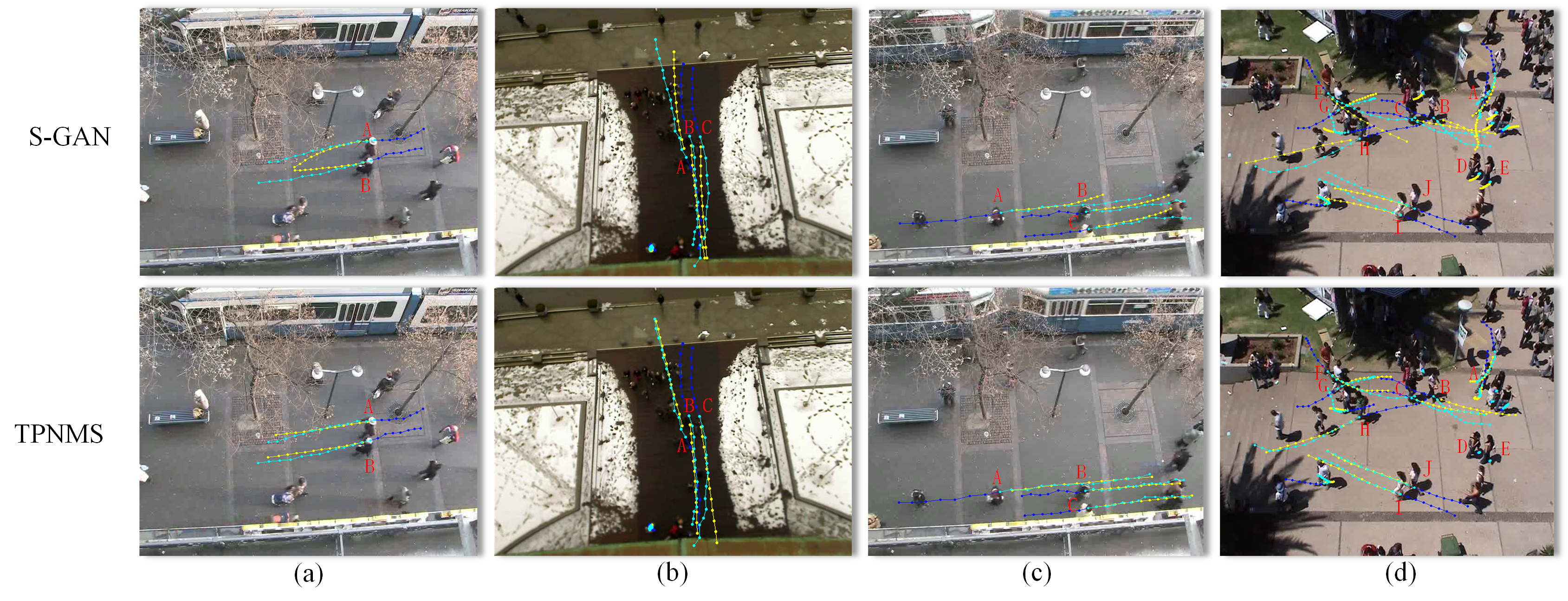}
	\caption{Examples of predicted trajectories by different methods. (a) walk in parallel; (b) meet from opposite directions; (c) follow people and (d) walk with complex social interactions. Blue line represents the historical trajectory; green line denotes the true future trajectory; yellow line shows the predicted future trajectory, and dots are the locations at different time steps. }
	\label{fig:qual_result}
\end{figure*}
\subsection{Quantitative Analysis}
Table \ref{tb:comparison} summarizes the results of different algorithms, where we report the average results for each method in the last two columns. We can see that all the algorithms perform much better than the linear model. Based on the results, we further draw the following conclusions:
\begin{itemize}
	\item Overall, our method TPNMS outperforms all the previous approaches in terms of the average ADE and FDE. 
	\item Compared with the baseline approach S-GAN \cite{gupta2018social}, TPNMS achieves significant performance gains. For example, S-GAN has an average error of 0.58 on ADE, and 1.18 on FDE, while TPNMS has much lower ADE (0.38) and FDE (0.73), corresponding to $34\%$ and $38\%$ relative improvements, respectively. This demonstrates that our proposed temporal pyramid network with multi-supervision indeed helps for pedestrian trajectory prediction.
	\item For the previous state-of-the-art method Social-STGCNN \cite{mohamed2020social}, TPNMS still achieves noticeable performance gains. For instance, TPNMS decreases the error of about $14\%$ on ADE and about $3\%$ on FDE compared with Social-STGCNN.
	\item Even without using any side information, TPNMS outperforms those methods utilizing the scene context, such as PIF \cite{liang2019peeking}, Sophie \cite{sadeghian2019sophie} and Social-BiGAT \cite{socialbigat2019}. This implies that the performance of TPNMS could potentially be improved by considering the scene context.
\end{itemize}

\begin{figure*}[!t]
	\centering
	\includegraphics[width=0.95\linewidth]{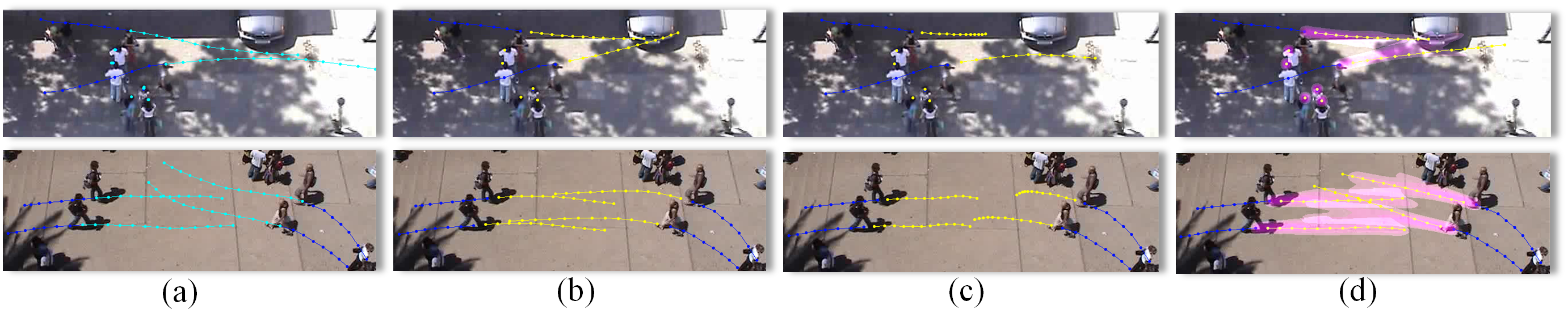}
	\caption{Examples of diverse predictions. The historical trajectory and predicted trajectory are marked in blue and yellow, respectively.
		 (a) The ground-truth future trajectory (green). (b, c) two examples of diverse predictions. (d) The density of the prediction, where the purple area is the visualization result of the predicted 20 pedestrian trajectories after mean filtering.}
	\label{fig:diverse}
\end{figure*}
\subsection{Qualitative Analysis}
In this subsection, we provide some examples to show how our TPNMS successfully captures complex motion behaviors of pedestrians.
We qualitatively compare the prediction results between Social-GAN and TPNMS.
\subsubsection{Results in different interaction scenarios}
We visualize examples from 4 scenarios in Figure \ref{fig:qual_result}.

\textbf{Walking in parallel } When people are walking side by side, they usually have tight connection to each other, and their relative positions tend to be preserved and motion behaviors tend to change consistently. In the Fig. \ref{fig:qual_result}(a), two target pedestrians A and B are walking in parallel.
It can be noticed that S-GAN incorrectly predicts that these two pedestrians will walk across each other, and have a high possibility of collision. 
Compared with S-GAN, the predictions by our TPNMS show that these two pedestrians will keep walking in parallel, which is close to the ground-truth trajectories marked by green lines.
This demonstrates the superiority of modeling motion behavior at various tempos.

\textbf{Meeting from opposite directions} People avoiding each other when moving in opposite direction is common in reality. Fig. \ref{fig:qual_result}(b) presents a scenario where two groups are meeting from opposite directions. We can see that the local behaviors of persons A, B and C are adjusted slightly to avoid collision. Compared with S-GAN, the trajectory of the person A predicted by TPNMS is more accurate after meeting. Further, TPNMS successfully predicts that persons B and C will keep walking in parallel, while the forecasts of S-GAN deviate from their true behaviors.

\textbf{Following people} When a person is following someone, he or she might want to draw attention to the person ahead, and maintain a safe distance between them. Fig. \ref{fig:qual_result}(c) shows a situation where the person A is walking behind the person B. We can see that S-GAN tends to decrease the speed of person A even when the distance between others is sufficiently large, while our proposed models TPNMS can more accurately forecast the speed at each time step, and still preserve a safe distance to avoid collision.

\textbf{Walking with complex social interactions} 
The complex interactions drive people using various ways to avoid collisions. As shown in Fig. \ref{fig:qual_result}(d),  many trajectories generated by S-GAN have large deviations from the ground-truth ones, e.g., the persons A, F, H and I. Besides, we can see that S-GAN fails to adjust the behaviors of persons A, B and C, and then causes a collision at the end of the predicted trajectories.  As for TPNMS, the predicted trajectories match better with the ground-truth one. For instance, persons I and J are maintained walking in parallel. Furthermore, we observe that the speed of person A is clearly slowed down by TPNMS, to avoid collision with persons B and C. For the trajectories of persons F and H, TPNMS achieves much better prediction accuracy than S-GAN. This demonstrates that our method can effectively capture the motion behaviors of pedestrians in scenarios of complex social interactions. 

\subsubsection{Results of diverse predictions} 
Our model is capable of producing multiple plausible and diverse trajectories conforming to the multimodal behavior of pedestrians.
In Fig. \ref{fig:diverse}, we show some examples of diverse predictions by sampling the noise vector $z$ from the standard normal distribution. 
We can see from Fig. \ref{fig:diverse}(b) and Fig. \ref{fig:diverse}(c) that our model generates two socially acceptable and distinct trajectories with different $z$, including changing the direction and speed. For instance, the top image of Fig. \ref{fig:diverse}(b) shows that the person is walking toward the car, where the direction is different from the true trajectory but the predicted path is still acceptable. Similar phenomenon can also be observed from the bottom image of Fig. \ref{fig:diverse}(b). Besides, images presented in Fig. \ref{fig:diverse}(c) show that $z$ can also affect the speed of pedestrians. In Fig. \ref{fig:diverse}(d), we draw the density of the predicted trajectory by 20 randomly generated samples. The purple area constructs a plausible area that each pedestrian may pass. The position of darker color indicates a higher probability that the person will pass through. Furthermore, in Fig. \ref{fig:diverse}(d), we also plot the best predicted trajectory from 20 samples for each scenario, and we can see that it closely matches the true trajectory shown in Fig. \ref{fig:diverse}(a).

\begin{table}[!t]
	\renewcommand{\arraystretch}{1.5}  
	\centering  
	\fontsize{9.0}{10.0}\selectfont
	\begin{tabular}{cccc}  
		\hline  
		\multirow{2}{*}{\diagbox{Models}{Modules}}&\multirow{2}{*}{TP}&\multirow{2}{*}{Multi-}&\multirow{2}{*}{AVG}\cr\ &Layer&Supervision&(ADE/FDE)\cr
		\hline
		S-GAN-P &$\times$&$\times$&0.61/1.21\cr 
		TPN&\checkmark&$\times$&0.41/0.79\cr 
		TPNMS&\checkmark&\checkmark&\bf0.38/0.73\cr
		\hline  
	\end{tabular}
	\caption{The average ADE/FDE performance of variants.}
	\label{table_2}  
\end{table}

\subsection{Ablation Experiments}
In Table \ref{table_2}, we systematically evaluate our method through a series of ablation experiments, where we consider the following variants of our method: \\
\textit{S-GAN-P}: the method without the temporal pyramid module and the multi-supervision module. With this setting, our method degrades to S-GAN-P; \\
\textit{TPN}: the method only considers the temporal pyramid module without the multi-supervision;\\
\textit{TPNMS}: the method considers both the temporal pyramid module and the multi-supervision.

Comparing TPN with S-GAN-P, we can see that TPN significantly reduces the ADE from $0.61$ to $0.41$, and the FDE from $1.21$ to $0.79$, which indicates that our temporal pyramid architecture can more effectively model the global context and local context of trajectories. 
Further, we observe that TPNSM can further improve the prediction accuracy in terms of ADE/FDE metrics, which demonstrates the importance of multi-supervision to ensure effective hierarchical representations.

\section{Conclusion}  \label{sec:conclusion}
In this paper, we have proposed a novel pyramid architecture for pedestrian trajectory prediction, which outperforms state-of-the-art methods on several benchmark datasets. First, we have devised a temporal pyramid network through squeeze and dilation modulations, which encodes and decodes the trajectory at multiple resolutions. This enables our method to capture both short-range and long-range motion behaviors of pedestrians. By resorting to a coarse-to-fine fusion strategy and the multi-supervision, our method can progressively merge high-scale global context with low-scale local context, finally resulting in an accurate trajectory prediction.
Finally, with a GAN based framework, our method can generate multiple socially-acceptable trajectories conditioned on the same trajectory history, obeying the multimodal property of pedestrians. Both quantitative and qualitative experimental results demonstrate the promising performance of our method under various situations.



\bibliography{TP}
\end{document}